\documentclass[preprint, 12pt]{amsart}
\usepackage[table]{xcolor}
\usepackage{times}
\usepackage{amssymb, amsmath}
\usepackage{amsthm}
\usepackage{tikz}
\usetikzlibrary{matrix,arrows}
\usepackage{enumitem}
\usepackage{booktabs}
\usepackage{color}
\usepackage{multirow}

\theoremstyle{plain}

\theoremstyle{remark}

\definecolor{Gray}{gray}{0.95}
\newcolumntype{g}{>{\columncolor{Gray}}c}
\graphicspath{{./images/}}

\calclayout

\DeclareUnicodeCharacter{202F}{\,}

\begin{document}
\title[ML applications in Covid]{Machine learning applications for COVID-19:\\ A state-of-the-art review.}

\author[F. Kamalov, A. Cherukuri, H. Sulieman, F. Thabtah, A. Hossain]{Firuz Kamalov$^1$$^{\boldsymbol{*}}$, Aswani Cherukuri$^2$, Hana Sulieman$^3$, Fadi Thabtah$^4$, Akbar Hossain$^5$}

\address{$^{1}$ Department of Electrical Engineering\\
Canadian University Dubai, Dubai, UAE.}
\email{\textcolor[rgb]{0.00,0.00,0.84}{firuz@cud.ac.ae}}

\address{$^{2}$ School of IT and Engineering\\
 Vellore Institute of Technology, 
 Vellore, India}
\email{\textcolor[rgb]{0.00,0.00,0.84}{cherukuri@acm.org}}

\address{$^{3}$ Department of Mathematics and Statistics\\
 American University of Sharjah, Sharjah, UAE}
\email{\textcolor[rgb]{0.00,0.00,0.84}{hsulieman@aus.edu}}

\address{$^{4}$ School of Digital Technologies, Manukau Institute of Technology\\
Auckland, New Zealand}
\email{\textcolor[rgb]{0.00,0.00,0.84}{fadi.fayez@manukau.ac.nz}}

\address{$^{3}$ School of Engineering, Auckland University of Technology\\
Auckland, New Zealand}
\email{\textcolor[rgb]{0.00,0.00,0.84}{akbar.hossain@manukau.ac.nz}}

\date{\today
\newline \indent $^{*}$ Corresponding author}

\begin{abstract}
The COVID-19 pandemic has galvanized the machine learning community to create new solutions that can help in the fight against the virus.  The body of literature related to applications of machine learning and artificial intelligence to COVID-19 is constantly growing. The goal of this article is to present the latest advances in machine learning research applied to COVID-19. We cover four major areas of research:  forecasting, medical diagnostics, drug development, and contact tracing. We review and analyze the most successful state of the art studies. In contrast to other existing surveys on the subject, our article presents a high level overview of the current research that is sufficiently detailed to provide an informed insight.
\end{abstract}

\maketitle


\section{Introduction}

The field of machine learning has made tremendous progress over the past decade. Improved  deep learning algorithms coupled with increased computational capacity catalyzed the growth of the field into stratosphere. As a result, machine learning has been used in a diverse array of applications. Arguably the most crucial application of machine learning has been in the fight against COVID-19 pandemic. Researchers have aggressively - and often successfully - pursued a number of different avenues using machine learning to battle COVID-19. A range of machine learning applications have been developed to tackle various issues related to the virus. In this paper, we present the latest results and achievements of the machine learning community in the battle against the global pandemic. In contrast, with other existing surveys on the subject we provide a general overview that is nuanced enough to provide a substantial insight. Our survey includes preprint works to ensure the most up-to-date coverage of the topics.
The current applications of machine learning to COVID-19 can be divided into four groups: 
\begin{itemize}
\item forecasting
\item medical diagnostics
\item drug development
\item contact tracing
\end{itemize}
Deep learning algorithms have been successfully deployed to forecast the number of new infections. Recurrent neural networks  have shown superior performance in time-series forecasting over traditional approaches such as ARIMA models. Researchers have used recurrent networks, and their variant long short-term memory networks, to successfully model the spread of the infection and predict the future number of infections in population.
Arguably the most important application of machine learning is in the field of medical diagnostics that is made possible by the advances in computer vision.
Machine learning has achieved near human level accuracy in many image recognition tasks. Therefore, it is no surprise that   image recognition software is successfully being used to detect signs of COVID-19 in patient chest X-ray images. In many parts of the world where an effective clinical testing procedure is not available or unaffordable chest X-ray images and CT scans provide the only option to diagnose the virus. Studies have shown that deep leaning approaches can diagnose COVID-19 based on chest X-ray image with over 99\% accuracy. 
Smart contact tracing using artificial intelligence  has helped authorities locate potential infected persons. A number of software solutions based on artificial intelligence are currently in use to trace spread of the virus. 
Machine learning has been used to help guide researchers to new discoveries in pharmacology. In particular, variational autoencoders have the ability to analyze perturbations in chemical composition that can lead to possible new medicines. Applying autoencoders to the existing flu vaccines can help identify potential avenues to creating COVID-19 vaccine.

The challenge to fight off the global pandemic and help the humanity has spurred researchers across disciplines. In an effort to accelerate scientific research on COVID-19 the publishing community  has made all the related publications freely available to the public. As a result, we are able to access and assess all the current research and present our survey to the readers. Our goal is to provide a quick, but sufficiently detailed, overview  of the current state of the art in machine learning research applied to COVID-19. We hope our survey will supply the reader with the necessary information to facilitate a deeper investigation into the topic.

The paper is structured as follows. In Section 2, we discuss the use of machine learning in forecasting the number of new infections. Section 3 discusses the use of deep learning in detection and diagnosis of the infection. Section 4 contains the information about the use of machine learning in drug discovery and development. Section 5 discusses the current research related to the application of machine learning for contact tracing. Finally, Section 6 concludes the paper with a few closing remarks.


\section{Forecasting}
Forecasting the number of infections is critical for proper planning and allocation of resources. Modern machine learning (ML) algorithms such as long short-term memory (LSTM) networks have been shown to outperform the traditional time series models such ARIMA  and GARCH. As a result, LSTMs have been used in various application involving time series projections \cite{Gurrib, Kamalov}. Several countries employ ML based software to estimate the number of future infections and the trajectory of the infected population. In this subsection, we will provide an overview of the latest advances in ML related to forecasting the number of COVID-19 infections. The results of our survey are summarized in Table \ref{forecasting} and a more detailed discussion follows below.

A comparative study of ML-based algorithms for COVID-19 forecasting was done in \cite{Ardabili}. The authors analyzed a number of evolutionary algorithms such as Genetic Algorithm, Particle Swarm Optimization, and Gray Wolf Optimizer as well as  ML algorithms such Multilayer Perceptron (MLP) and adaptive network-based fuzzy inference system (ANFIS). The models were evaluated on the basis of their accuracy for different prediction lead times. The authors employed data from 5 different countries in their study experiments revealed that MLP and ANFIS algorithms produce the best results achieving correlation level of 0.999.
A novel approach from Google Research that combines temporal and spatial data is proposed in \cite{Kapoor}. Using graph neural networks and Google mobility data the authors uncover the rich interactions between time and space that is often present in the spread of pandemic. Numerical experiments demonstrate the power of mobility data with the GNN framework.
In \cite{Melin}, the authors  employ an ensemble neural network to predict the number of confirmed cases and deaths in Mexico. The proposed ensemble network (MNNF) consists of 3 modules: nonlinear autoregressive and function fitting neural networks. The module predictions are combined via a fuzzy integrator - designed to handle uncertainty - into a single output. The method is tested on data from Mexico. The authors carried out experiments to predict the number of confirmed cases and deaths 10 days ahead. Results reveal that the MNNF method outperforms single neural network models. 
The authors in \cite{Arora} test 3 LSTM-based models to forecast the number of infected individuals for 32 states in India. The tested models include stacked, convolutional, and bi-directional LSTM neural networks. The predictions are made one day and one week ahead. The results show that the bi-directional LSTM produces the optimal results.
Several ML models are compared in \cite{Silva} to forecast confirmed cases in Brazil and the US. The models under consideration include Bayesian neural network, cubist regression, kNN, random forest, and SVR. In addition, variational mode decomposition (VMD) is applied as a preprocessing step. The authors also consider exogenous variables such as temperature and precipitation. Numerical experiments produce mixed results with no clear  favorite. It can only be noted that VMD improves model performance when the prediction horizon is 6 days ahead.
The authors in \cite{Shahid} compare statistical and ML approaches to time series forecasting. In particular, they study autoregressive integrated moving average (ARIMA), support vector regression (SVR), and LSTM models to forecast the number of infections, deaths, and recoveries. The model input consists of the data from the previous 110 days. The model is used to predict infections for the next 48 days. The study is based on data from 10 countries. The results show that LSTM models generally outperform ARIMA and SVR. 
Machine learning approaches do not always outperform traditional methods. In \cite{Rustam}, the authors compare classic statistical methods to SVR to predict the number of positive cases, death rate, and recovery rate. The study covers a large number of countries. Results show that statistical models outperform SVR.
In \cite{Direkoglu}, the authors apply deep learning to forecast the number of infections and deaths regionally and worldwide. LSTM models use observed last 3 days of data to forecast 10 days ahead. In their analysis the authors considered Middle East, Europe, China, and worldwide data. The results show that the forecasts achieve 1.5\% root mean square error (RMSE).

\begin{table}[!htbp]
\centering
\caption{ML based research studies in COVID-19 forecasting.}
\label{forecasting}
\begin{tabular}{|p{0.10\linewidth}|p{0.15\linewidth}|p{0.22\linewidth}|p{0.23\linewidth}|p{0.23\linewidth}|}
\hline
\sc Study &  \sc  Objective &    \sc Methodology  &  \sc    Data &   \sc Results \\
\hline
\scriptsize Kapoor et al., (2020) &    
\scriptsize Forecast daily new cases in the US. &    
\scriptsize       A spacio-temporal graph neural network is used to learn the complex interactions between the time and mobility data. The model is implemented via GNN framework. 
 &  \scriptsize NYT COVID-19 dataset and the Google Mobility Dataset/Reports. The data includes positive cases in the US  over the period of Jan-Apr, 2020.
 & \scriptsize GNN model achieves correlation of 0.998.\\ \hline

\scriptsize Ardabili et al., (2020) &    
\scriptsize Comparative study of ML forecasting models. &   
\scriptsize       Compared GA, PSO, GWO ANFIS, and MLP approaches  and their accuracy for different lead-times.

 &  \scriptsize Worldometer. Number of COVID-19 cases for 5 countries over the period of Jan-Mar, 2020.
 & \scriptsize ANFIS and MLP produce the best results with correlation of 0.999.\\ \hline

\scriptsize Melin et al., (2020) &     
\scriptsize Make up to 10-day ahead predictions of number of confirmed cases and deaths. &     
\scriptsize Ensemble of neural networks consististing of 3 modules: 2 nonlinear autoregressive networks (NAR) with different parameters and 1 function fitting network (FITNET). The output of the modules is combined via a fuzzy integrator. 
&  \scriptsize   Mexico Government (coronavirus.gob.mx). Confirmed cases of COVID-19 and related death cases in Mexico. The data on the state and country level over a 110-day period.
 &  \scriptsize \%RMSE predicted confirmed cases on the country level 0.0808; \%RMSE for the states ranges from 0.0322 to 0.2157. \%RMSE predicted number of deaths on the country level 0.0914; \%RMSE for the states ranges from 0.0175 to 0.2094.\\ \hline

\scriptsize Arora et al., (2020) &     
\scriptsize Predict the number of next day (and week) positive cases. &         
\scriptsize       Three LSTM-based models - stacked, convolutional, and bi-directional LSTM - are tested.
 &  \scriptsize Ministry of Health and Family Welfare (India). Number of postive cases in 32 states and territories in India over the period of Mar 14, 2020 to May 14, 2020.
 &  \scriptsize Average MAPE for stacked, convolutional, and bi-directional LSTM model are 5.05\%, 4.81\%, and 3.22\% respectively. \\ \hline

\scriptsize da Silva et al., (2020) &      
\scriptsize  Predict the number of new cases 1, 3, and 6 days ahead. &   
\scriptsize     Various ML models - Bayesian neural network, cubist regression, kNN, random forest, and SVR - are considered. In addition, variational mode decomposition (VMD) preprocessing is applied. Exogenous input variables - temperature and precipitation - are also considered. 
 &   \scriptsize  John Hopkins University repository and Brazilian State Health Offices API. Number of daily positive cases for 5 states in the US and Brazil until Apr 28, 2020.
& \scriptsize  Mixed results with different models achieving the best outcomes on various subsets of data. The best models achieved an out of sample forecasting error of 3\%. \\ \hline

\scriptsize Shahid et al., (2020) &    
\scriptsize Predict the number of positive, death, and recovery cases & 
\scriptsize       ARIMA, SVR, and LSTM models are compared. The models are applied to data from 10 countries.
 &  \scriptsize China Data Lab, 2020, "World COVID-19 Daily Cases." Number of confirmed, death, and recovery cases  over the period of Jan 22, 2020 to Jun 27, 2020. 
 & \scriptsize Mixed results, but in general LSTM appears to produce better results. For instance, LSTM has the lowest MAE values for confirmed cases and deaths as 2.0463 and 0.0095 respectively. \\ \hline

\scriptsize Rustam et al., (2020) &    
\scriptsize Predictions for the next 10 days based on the on data from previous 56 days. &    
\scriptsize       Linear Regression, LASSO, Support Vector Regression, and Exponential Smoothing (ES) are used. The predictions are made for infection, death, and recovery rates. 
 &  \scriptsize John Hopkins University repository. Number of confirmed, death, and recovery cases worldwide over the period of Jan 22, 2020 to Mar 27, 2020.
 & \scriptsize ES achieves the best results in death rate prediction with $R^2=0.98$. LASSO achieves the best results in confirmed rate prediction with   $R^2=0.98$. \\\hline

\scriptsize Direkoglu et al., (2020) &  
\scriptsize Use previous 3 days to forecast the next 10 days &      
\scriptsize       LSTM model is used to forecast the number of new cases and deaths regionally and worldwide. 
 &  \scriptsize WHO, CCDCP, and Worldometer. Number of confirmed, death, and recovery cases worldwide over the period until Apr 10, 2020.
 & \scriptsize The trained LSTM model achieves 1.5\% RMSE. \\

\hline
\end{tabular}
\end{table}

\section{Medical diagnostics}
Diagnosing COVID-19 infection is a key first step to fighting the virus. The rapid spread of the disease across the globe has made diagnosis of the disease at early stages not only important for the individual patient but also for preventing the community spread of the disease.
Polymerase chain reaction (PCR) tests that are currently employed to detect the presence of the COVID-19 virus require time and capital to administer. Despite recent improvements PCR tests remain scarce and costly in developing countries and rural areas. PCR tests may further suffer from sample preparation and quality control which can lead to insufficient sensitivity \cite{Shi}. Therefore, developing alternative approaches to testing is a vital research area. At present there are several ML applications that support diagnostic process.  Deep neural networks demonstrated capability to achieve high accuracy in image detection tasks. Consequently, applying deep learning and other ML techniques to X-ray and CT scan images has been one of the intensely researched areas. In addition, detection approaches based on clinical data have also been tried and tested.  Artificial intelligence (AI) based methods augment the diagnosis process and accelerate the treatment of the disease.  These models can assist the physicians and healthcare professionals not only during testing and treatment but also for planning and managing of the resource \cite{Lalmuanawma}.   The results of our survey on the current AI/ML research for COVID-19 diagnostics are summarized in Tables \ref{diagnostics1} and \ref{diagnostics2}.

Imaging techniques such as X-rays and CT scans are widely used as diagnosis tools for many lung diseases including tuberculosis, lung cancer, and pneumonia viruses. CT scan images provide fast and detailed information about the pathology and prognosis of diseases.  As a result, ML techniques are being increasingly integrated with imaging and computer vision methods for applications in disease diagnosis.   
The success of deep learning techniques in detecting and diagnosing various types of pneumonia has been already reported in the literature. The authors in \cite{Li} developed a robust model based on 3-dimensional convolutional neural network (CNN) framework to extract features from CT scan images and distinguish COVID-19 from the community acquired pneumonia.  When diagnosing patients in early stages, AI models proved to be successful by integrating both CT scan imaging and clinical information \cite{Mei}.  Combining the output of CNN model on CT scan images and the output of ML models such as SVM and Random Forests on clinical data the accuracy of diagnosis reaches the levels of human healthcare experts.  CT scan imaging is the diagnostic tool predominantly used in treating the pulmonary infections.  The same is employed during the current outbreak by many countries in diagnosing COVID-19 patients - particularly at early stages.  Further progress was made by Zhou et al. \cite{Zhou} who identified the importance of segmentation and proposed deep learning based models to address these issues in ML based diagnosis of COVID-19.

\begin{table}[!htbp]
\centering
\caption{ML based research studies in COVID-19 medical diagnostics.}
\label{diagnostics1}
\begin{tabular}{|p{0.10\linewidth}|p{0.15\linewidth}|p{0.22\linewidth}|p{0.23\linewidth}|p{0.23\linewidth}|}
\hline
\sc Study &   \sc Objective &  \sc   Methodology  &   \sc   Data &  \sc Results \\
\hline

\scriptsize Zhang et al., (2020)  &      
\scriptsize  Diagnose COVID-19 based on CT scans. &   
\scriptsize    The proposed system consists of two segmentation models: one for lung lesion segmentation and another one for diagnosis prediction.   3D convolutional blocks are used for the classification. 
 &   \scriptsize  A large database of 532,506 CT scan images of 4,154 patients having COVID-19, common pneumonia and normal controls.   
& \scriptsize  Results are evaluated over the dice coefficient and pixel accuracy metrics.  The system achieved the accuracy of 92.49\%, sensitivity of  94.93\% and 91.13\% specificity. \\ \hline

\scriptsize Apostolopoulos et al.,  (2020)  &    
\scriptsize Analyze the adoptability of CNN techniques for diagnosis of COVID-19 with the help of X-ray images.    & 
\scriptsize       Transfer learning with CNN is used over the X-ray image data.  
 & \scriptsize An image dataset of 1427 X ray images that include COVID-19 positive, common pneumonia and normal conditions. 
  & \scriptsize The model achieved specificity of 96.46\%, sensitivity of 98.66\%, and accuracy of 96.78\%.   Results vary according to the severity of symptoms. \\ \hline
  
 \scriptsize Zheng et al., (2020)  &  
\scriptsize Detect COVID-19 via deep learning. &      
\scriptsize       Deep CNN network with three stages: first stage with 3D convolution, batchnorm and pooling layers; second stage has two 3D residual blocks; third stage is with progressive classifier which abstracts the information by 3D max pooling and output the probability of COVID-19. CT scan images were pre-processed by a simple 2D UNet to create 3D lung mask.   
 &  \scriptsize 540 patients chest CT scans are considered for evaluation.  Further, travel and contact history of these patients, symptoms, clinical lab findings are also considered. 
 & \scriptsize The proposed model achieved sensitivity of 90.7\%, specificity of 91.1\% and AUC  of 97.5\%.    \\\hline
 
 \scriptsize Sun et al., (2020)   &  
\scriptsize Detect and classify the COVID-19 from CT scan images of chest. &      
\scriptsize       Deep forest algorithm based on adaptive feature selection.  From the CT scan images, location specific features are extracted.    The trained deep forest model is used to reduce the redundancy of features and classify COVID-19 and community acquired pneumonia.
 &  \scriptsize A dataset of 2522 chest scan images are considered
 & \scriptsize The proposed model achieved sensitivity of 93.05\%, specificity of 89.95\%, accuracy  of 96.35\%.  \\\hline
 
 \scriptsize Kasani et al., (2020)  &  
\scriptsize Automatically detect the COVID-19 in X-ray and CT Scan images. &      
\scriptsize       Compared the performance of a pool of deep learning based different feature extraction models such as DenseNet, MobileNet, ResNet, InceptionV3, NASNet.  Features extracted from these models are fed to classification techniques such as decision trees, random forests, XGBoost, AdaBoost, Bagging classifier, and LightGBM.    
 &  \scriptsize A total of 137 images among which 117 are X-ray and 20 are chest CT scans images.  These are COVID-19 positive patients data and similar in number healthy patients data. 
 & \scriptsize PMobileNet and Inception V3 with Bagging classifier have provided the best classification performance with 99\% for precision, recall and F Score.  However the data is limited.    \\\hline

  \scriptsize Li et al., (2020)  &    
\scriptsize Develop automatic ML framework to detect coronavirus with the help of chest CT scan images. &    
\scriptsize       Three dimensional deep learning framework aimed to extract two dimensional local features and 3 dimensional global features.  Then the features are submitted to a fully connected network with a softmax activation
 &  \scriptsize T4352 CT scans collected from 3322 patients. Among these, 30\% were for Coronavirus, 40\% for community acquired pneumonia, and remaining for non-pneumonia abnormalities.
 & \scriptsize High specificity and sensitivity is achieved. \\ \hline

\end{tabular}
\end{table}


\begin{table}[!htbp]
\centering
\caption{ML based research studies in COVID-19 medical diagnostics (cont).}
\label{diagnostics2}
\begin{tabular}{|p{0.10\linewidth}|p{0.15\linewidth}|p{0.22\linewidth}|p{0.23\linewidth}|p{0.23\linewidth}|}
\hline
\sc Study &  \sc  Objective & \sc    Methodology  &   \sc   Data &   \sc Results \\
\hline

\scriptsize Nour et al., (2020)  &    
\scriptsize Detect the positive cases of COVID-19 automatically from chest X ray images. &    
\scriptsize       A serial five convolutions layer network is designed to learn the model from the scratch.  Optimization of the model parameters are made through Bayesian optimization. Using the learned model, deep feature vectors are extracted.   For classification, SVM based on RBF kernel, C4.5 decision tree and k-nearest neighbour techniques are adopted.  
 &  \scriptsize An open access X ray image database containing 2905 images are considered.  The images are belong to three classes: COVID-19, normal and viral pneumonia.    
 & \scriptsize The proposed model with SVM classification achieved accuracy of 98.75\%, sensitivity of 89.39\%, specificity of 99.75\% and F-score of 95.75\%.  The main concern is that if X-ray images are relevant for early diagnosis. \\\hline

\scriptsize Wang et al., (2020)  &  
\scriptsize Perform diagnosis and prognosis of COVID-19 using CT. &      
\scriptsize       Proposed a DenseNet like structure with multiple stacks of convolution, ReLu activation, max pooling and batch normalization operations.   
 &  \scriptsize CT scan images of 5372 patients among which 4106 patients data is used to pre-train the model and 1266 patients data to test and validate. Prognostics analysis is made with 471 patients who are tested COVID-19 positive are considered. While considering the ROI, 3D bounding box of entire lung is considered. 
 & \scriptsize Proposed model exhibited good performance with AUC of 90\%, specificity of 89.93\% and sensitivity of 78.93\%.   The prognosis should be based on the severity of the symptoms.   \\\hline

\scriptsize Mei et al., (2020)  &    
\scriptsize Design an AI model that can diagnose the COVID-19 patients in the early stage by integrating CT scans, clinical symptoms, contact history and testing. &   
\scriptsize       CNN’s were deployed in two layers. The first CNN model is slice selection CNN that select the abnormal lung scans.  The second CNN is to diagnose and predict the likelihood of COVID.  Further, patient details were fed to ML model such as SVM/ MLP/RF to classify COVID-19 positivity.  The third output was obtained by integrating result of diagnosis CNN model and patient’s clinical information to a MLP. 
 &  \scriptsize CT Scan image dataset of 905 patients.  Further, clinical information details of these patients are also considered.  
 & \scriptsize Results of the proposed AI model that combined both CT images and clinical information during diagnosis are equivalent to the accuracy of human experts particularly for early stage cases.\\ \hline

\scriptsize Burdick et al., (2020)  &     
\scriptsize Predict the need and requirements of ventilation during the diagnosis of COVID-19 patients.  &     
\scriptsize XGBoost classifier based method which uses an ensemble learning technique  to learn and classify the low risk or high risk category patients.     
&  \scriptsize   197 patients with positive diagnosis of COVID-19.  
 &  \scriptsize The specificity and sensitivity results indicate that the model was able to identify and discriminate the patients who require the ventilation support.  \\ \hline

\scriptsize Elaziz et al., (2020)  &     
\scriptsize Detect COVID-19 infection based on X-ray image. &         
\scriptsize       Feature extraction based on Fractional Multichannel Exponent Moments and feature selection based on modified Manta-Ray Foraging Optimization based on differential evolution are implemented in a multicore environment.
 &  \scriptsize 216 positive cases and 1675 negative cases of COVID-19.  Another dataset with 219 and 1,341 positive and negative cases.   
 &  \scriptsize Significant gain in computational time.  Accuracy of the method is proved to be better than other optimization methods. \\ \hline

\end{tabular}
\end{table}

 
Despite the promising research results there is still a lot of room for growth for ML based diagnostics. Production ready applications that can be used in hospitals require further refinement.
A great deal of of research is yet to be conducted to improve their reliability.  The main challenge in deploying the AI/ML models in the COVID-19 is the generalization ability of these models which is also prevalent in AI based models in other applications. Another major bottleneck in implementing AL/ML based solutions in healthcare is the availability of patient data samples of necessary size and quality to train the ML models.  In some instances though the data is available, format and structure of the data pose another challenge.  Integrating existing research solutions to practical applications and products is another challenge.  Finally it is vital to ensure that the studies, investigations conducted and reported during this pandemic and pressing times are technically, scientifically and ethically are correct.   

A wide array of ML models has been deployed to try diagnose instances of COVID-19. The list of models includes CNN, RNN, SVM, transfer learning, XGBoost and others. Although these models demonstrate high performance and accuracy they possess limitations such as the lack of sufficient data to train the models, inability to generalize the results, etc \cite{Zhou}.
Despite the ongoing efforts to apply ML/AI in COVID-19  diagnostics some members of the radiologist community have raised their concerns regarding possible pitfalls.
Laghi \cite{Laghi} has cautioned that while AI/ML should be used for diagnosis of COVID-19, a more objective and precise quantification is required in understanding the lungs involvement of disease.  Wynants \cite{Wynants}  reviewed the validity and usefulness of the various models published in the literature on COVID-19 diagnosis, prognosis and risk prediction.  Their analysis over 145 models in 107 published documents showed that there exists a high risk of bias. The results of these models are probabilistic and hence are not recommended to be adopted for practical use.  They call for more rigorous analysis of these models with proper methodological guidance and provision of description of populations under study.  They  also warned that if the studies are unreliable, it would lead to harmful effects in diagnosis and prognosis of the disease.  
Based on the careful review of the existing literature on ML based diagnostics for COVID-19 we conclude that the proposed models have significant potential. The existing models can be used as stepping stones for building more robust and resilient models that would assist the healthcare professionals in diagnosis and decision making.  AI/ML researchers should learn from the experiences of this pandemic and focus on developing  models in collaboration with healthcare professionals and medical experts. We note that the most important challenge is the availability of data to train the models as well as the treatment of the data.  Resolving this issue can have a big impact on the robustness, generalization ability of models for practical applications.


\section{Drug development}
Machine learning algorithms are increasingly being used to search for new chemical combinations that can lead to effective medicine. Artificial intelligence and machine learning techniques have become an integral part of the pharmaceutical world. Integrating these techniques into the complex drug developing pipeline has proven to be both cost-effective and less time-consuming. Machine learning techniques are particularly useful as they provide a set of tools that improve the process of drug discovery and development for specific situations with the help of available data that is reliable and of high quality. As a results a large effort has been under way to apply AI/ML based solutions in pharmacology. A summary of the survey of the current literature in the field is provided in Table \ref{pharma1}. Several pharmaceutical companies have employed ML-based algorithms such as artificial neural networks, Support Vector Machines (SVM), deep learning and many others  to develop various drugs and vaccines \cite{reda20}. The authors in \cite{reda20} provide a review of recently developed algorithms to design automated drug development pipelines consisting of drug discovery, drug testing and drug re-purposing. In drug discovery, the deep learning algorithm Generative Adversarial Networks (GAN) is used to identify DNA sequences associated with specific functions and Bayesian Optimization (BO) is used to produce proteins of interest with lower costs.   In drug testing, sequential decision-making algorithms such as the Bayesian-based Multi-Armed Bandit (MAB) algorithms  are used to test several drug candidates and determine the best treatments.   In drug re-purposing, text mining methods and graph-based recommender systems are used to identify correlations and predict drug-disease interactions. The authors compiled a list of relevant  data sets  for drug development pipeline studies.

 In 2019, the National
Institute of Allergy and Infectious Diseases sponsored the first U.S. clinical trial to develop a vaccine against SARS-CoV-2 using an AI-based model \cite{Ahuja20}. An AI program called synthetic chemist was created to generate trillions of synthetic compounds and another AI-based program called Search Algorithm for Ligands (SAM) was used to sift through the trillions of compounds and determine the most suitable candidates as vaccine adjuvants. With the fast spread of COVID-19, there has recently been a race in utilizing ML techniques and AI capabilities to develop an effective vaccine and antivirals.
The authors in \cite{abbasi20}  incorporated reverse vaccinology, bioinformatics, immunoinformatics and deep learning strategies to build a computational framework for identifying probable vaccine candidates and constructing  an  epitope-based  vaccine against COVID-19. The screening of viral proteome sequences resulted in short listing of Spike protein or Surface Glycoprotein of SARS-CoV-2 as a potential protein target that can be used to design the vaccine. The  physicochemical  properties of the protein were further examined using LSTMs
 and the results showed that the protein  is the primary responsible  for the  pathophysiology of SARS-CoV-2. The authors proposed that their computational pipeline can be used to design effective and safe vaccine against COVID-19.
In \cite{Yazdani20}, the authors used an 'In-Silico' analysis to design a potent multi-epitope peptide vaccine against SARS-CoV-2. MLP and SVM algorithms were used to screen for potential epitopes. The vaccine immunogenicity was enhanced using  three potent adjuvants and its tertiary structure was predicted, refined and validated using appropriate strategies. The results showed that the vaccine can interact effectively with toll-like receptors (TLR) 3, 5, 8 and by using in silico cloning, it has demonstrated a high-quality structure, high stability and potential for expression in Escherichia coli.
The authors  in \cite{ong20}  surveyed existing literature about COVID-19 and vaccine development. They used  Vaxign Reserve Vaccinology (VRV) tool and  Vaxign-ML,  a machine learning-based vaccine candidate prediction and analysis system, to predict and evaluate potential vaccine candidates for COVID-19. The results showed that in addition to the commonly used S protein, the non-structural protein (nsp3) was found to be second highest in protective antigenicity. Further investigation of the the sequence conservation and immunogenicity of the multi-domain nsp3 protein, the authors concluded that the nsp3 can be an effective and safe vaccine target against COVID19.

For the development of drug treatment for COVID19, the authors in   \cite{beck20} used a pre-trained deep learning-based drug-target interaction model called Molecule Transformer-Drug Target Interaction (MT-DTI) to predict any commercially available antiviral drugs that could be effective against SARS-CoV-2. The model was compared to CNN-based model called DeepDTA and another two traditional machine learning based algorithms, gradient boosting and regularized least-squares model, using various data set. The MT-DTI showed the best performance in predicting the drug–target interactions and was able
 to identify various antiviral drugs such as redeliver, dolutegravir, efavirenz and atazanavir which could potentially be used in the treatment of SARS-CoV-2 infection.
 In \cite{Ke20}, the authors used deep neural networks (DNN) and established an AI platform to identify potential old drugs that could be used against the SARS-CoV-2. Different learning data sets consisting of compounds reported or proven active against SARS-CoV, SARS-CoV-2, human immunodeficiency virus (HIV), and influenza virus were generated and used to predict drugs
potentially active against coronavirusout of the marketed drugs. The predicted drugs were then tested and verified to serve as feedbacks to the AI platform for relearning and thus to generate a modified AI model. The implemented  AI-based framework was able to identify eight drugs with activities against Feline Infectious Peritonitis (FIP) coronavirus. The authors suggested that with prior use experiences in patients, these identified old drugs can potentially be proven to have anti-SARS-CoV-2 activity and hence be applied for fighting COVID-19 pandemic.
The authors in \cite{Kow20} analyzed over 10 million compounds using a machine learning pipeline in order to predict chemicals that interfere with SARS-CoV-2 targets. The pipeline  involves selection of
important physicochemical features for each target using recursive
feature elimination algorithms, followed by fitting
aggregated multiple support vector machines (SVM) models and regularized
random forest algorithm (regRF) to improve generalizability and then evaluating model performance using various computational validation methods. The authors concluded that their identified chemicals can accelerate testing of short-term and long-term  treatment strategies for COVID19.
 The importance of AI and Machine learning (ML) techniques that can accelerate 
 the discovery of a possible cure for COVID-19 is discussed in a recent review article by \cite{arshadi20}. The review article by  \cite{kan20} focused on the recent advances of
COVID-19 drug and vaccine development using artificial intelligence and discussed the potential of
intelligent training for the discovery of COVID-19 therapeutics.

\begin{table}[!htbp]
\centering
\caption{ML based research studies in pharmacology.}
\label{pharma1}
\begin{tabular}{|p{0.10\linewidth}|p{0.15\linewidth}|p{0.22\linewidth}|p{0.23\linewidth}|p{0.23\linewidth}|}
\hline
\sc Study & \sc   Objective &  \sc   Methodology  &  \sc    Data & \sc  Results \\
\hline

\scriptsize Ahuja et al., (2020)   &    
\scriptsize  Vaccine development against  COVID-19. &    
\scriptsize      Deep Neural Network-based  AI programs:  synthetic chemist and Search Algorithm for Ligands (SAM) are used.   
 &  \scriptsize COVID-19 Open Research Dataset (CORD-19). 
 & \scriptsize New approaches combining integrative medicine with AI models are emerging in the recent race to utilize ML techniques and AI capabilities to develop an effective vaccine against COVID-19. \\\hline

\scriptsize Abbasi et al., (2020)   &  
\scriptsize Identify a suitable vaccine candidate and construct an epitope-based vaccine against COVID-19. &      
\scriptsize       A computational framework using reverse vaccinology, bioinformatics, immunoinformatics and AI based strategies.  
 &  \scriptsize NCBI database in FASTA format. 
Crystal structures of human alleles: Protein Data Bank.
 & \scriptsize Spike protein (Surface glycoprotein), B-cell and T-cell epitopes were predicted for the construction of an epitope-based vaccine.   \\\hline
 
 \scriptsize Yazdani et al., (2020)   &  
\scriptsize Develop a multi-epitope peptide vaccine against the SARS-Cov-2. &      
\scriptsize       A variety of  ML methods including ANN, SVM, BepiPred, ANTIGENpro and VaxiJen are employed.
 &  \scriptsize NCBI database in FASTA format IEBD database. 
 & \scriptsize Designed-vaccine construct consists of several immunodominant epitopes has high antigenic capacity and induce humoral and cellular immune responses against SARS-CoV-2.    \\\hline
 
 \scriptsize Ong et al., (2020)    &  
\scriptsize Predict proteins candidate for vaccine against COVID-19. &      
\scriptsize       Vaxign- RV and Vaxign-ML strategies are used. The sequence conservation and immunogenicity of the predicted protein were further investigated.   
 &  \scriptsize NCBI and UniProt  databases. 
 & \scriptsize S protein had highest protective antigenicity score. nsp3 protein - with the second highest antigenicity score -was predicted as an alternate vaccine candidate.  \\\hline
 
 \scriptsize Beck et al., (2020)   &  
\scriptsize Predict binding affinity values between commercially available antiviral drugs and target proteins. &      
\scriptsize       Deep learning-based MT-DTI  was compared to  DeepDTA,   gradient boosting and regularized least-squares model.
 &  \scriptsize DrugBank, Drug Target Common (DTC) database, BindingDB  and  NCBI databases.   
 & \scriptsize MT-DTI performance best compared to a DeepDTA  and ML-based algorithms(SimBoost and KronRLS). 
Atazanavir is the best chemical compound against the SARS-CoV-2 3C-like proteinase.     \\\hline

\scriptsize Ke et al., (2020)     &    
\scriptsize  Identify pre-existing drugs with anticorona virus activities. &    
\scriptsize     Deep learning-based  AI-system was established. The predicted drugs were tested in vitro for verification and the results were fed back to the AI system for relearning. 
 &  \scriptsize DrugBank:https://www.drugbank.ca 
Database 1: SARS-COV, HIV, Influenza, other drugs. 
Database 2:  210 inhibitors of the 3C-like protease of SARS-CoV. 
 & \scriptsize 8 marketed drugs: Vismodegib, Gemcitabine, Clofazimine, Celecoxib, Brequinar, Conivaptan, Bedaquiline and Volcapone were identified as potential candidates against COVID-19.  \\\hline

\scriptsize Kowalewski \& Ray, (2020)    &  
\scriptsize Identify chemicals that interfere with SARS-CoV-2 targets. &      
\scriptsize      A machine learning drug discovery pipeline consisting of multiple support vector machines (SVM) models and random forest algorithm (RFA) was developed
 &  \scriptsize NUNII, DrugBank, Therapeutic Targets and bioassay database, ZINC.
 & \scriptsize SVM model with the RBF kernel outperformed regularized Random Forest or performed comparably.    \\\hline
 
 \scriptsize Keshavarzi et al., (2020)     &  
\scriptsize Review of ML-based drug development. &      
\scriptsize      A multifaceted and comprehensive investigation of existing literature of AI-based approaches (SUMMIT, GAN, RF, SVM, RFE, LSTM) used for COVID-19 drug and vaccine development. The results were used to create a database: CoronaDB-AI. 
 &  \scriptsize Complete toxicity dataset from distinct databases, including ToxCast and Tox21. Comprehensive epitope-based dataset. 
 & \scriptsize SML-aided molecular docking is one of the most prevalent approaches for virtual screening. 
3CLpro is the most popular target for virtual screening. 
Spike protein has been the most popular candidate for virtual vaccine discovery  \\\hline
\end{tabular}
\end{table}


\section{Contact tracing}
Effective contact tracing is a major factor in a virus containment strategy \cite{Ahmed}. In conventional contact tracing, a health care professional interviews the infected patient to trace and discover other individuals who may potentially be infected though contact with the patient. The main challenge of the conventional approach is the difficulty for an individual to recall all his contacts. In addition, the process requires availability of specialized clinicians using their experience and other resources \cite{Dar}. Recent technological improvements allowed the contact tracing process to be optimized with less human intervention in an intelligent approach known as digital proximity (DP) contact tracing. The DP approach utilizes network technologies to identify and locate individuals who could be potentially infected through contact.  

With he widespread availability of computing networks and mobile applications - and their associated technologies including smartphone, smartwatch, and others - most of the technology-based contact tracing systems are built on mobile platforms \cite{Ardakani, Maghdid}. These systems, named digital contact tracing (DCT), enable a registered user’s exposure to be evaluated through wireless signals such as Bluetooth low energy. Alternative technology-based tracing systems that are non-mobile and application-based utilize tracking information collected from a variety of sources such as banking transactions, security camera footage, GPS data from vehicles, mobile phones and others to estimate the proximity of an individual to an infected person.  

Artificial intelligence and machine learning - in particular deep learning algorithms - have been successfully used in medical diagnosis and screening systems due to their exceptional learning capabilities. In the context of DCT systems, these technologies can be incorporated to aid the decision-making process and improve the detection accuracy of contact tracing. Concretely, the data collected from registered users such as their daily tracks and geo locations in the DCT system are explored by the ML algorithm within digital platforms to provide medical professionals and government officials with useful insights.  Artificial intelligence and machine learning applications are currently utilized through the entire life cycle of COVID-19 starting from detection  to mitigation \cite{Lalmuanawma}. In contact tracing, a virtual AI agent is an alternative to a health professional in the case of classical contact tracing. The virtual AI agent with natural language capabilities can collect the information previously gathered by a health professional. In DCT systems, Bluetooth technology is  widely employed as a proximity detector for COVID cases. However, the performance of Bluetooth-based contact tracing apps may be affected by changing signal intensity, which can be exhibited by different mobile devices, mobile positions, body positions, and physical barriers \cite{Zhao}. Generic wireless multipath effects and shadowing are persistent issues which can lead to false positive and false negative identification. To improve the proximity detection accuracy in DCT systems, ML techniques can be used to analyze the Bluetooth signal and other phone sensors’ data.  

Recently, a 2-stage classifier was proposed that utilizes vanilla neural network to extract features from a signal emanating from different sources \cite{He}. Employing a deep learning technique directly on a smartphone involves high computational cost and power consumption. Therefore, during the first stage raw data from different sources is converted into fixed-length vectors and stored in the database. In the second stage, the vanilla deep learning algorithm is applied to detect proximity \cite{He}. A similar project under the TC4TL challenge compares several deep learning models including Conv 1d \cite{Lwowski}, support vector machines \cite{Shubina}, and decision tree-based algorithms \cite{Eva} to evaluate the accuracy of Bluetooth-based distance measurement \cite{Shankar, TC4TL}. The performance of different techniques is measured based the lowest normalized decision cost function (NDCF) which represents proximity detection performance considering the combination of false negatives and false positives.  
The results show that the Conv 1d network has the lowest NDCF.

It is evident that the performance of classification algorithms varies widely based on proximity thresholds. For example, Song \cite{Song} reported that when considering two people six feet apart in classifying Bluetooth beacon RSSI values, a Gaussian support vector machine classifier yielded better accuracy than a decision tree classifier. For validation, each experiment was conducted by placing two Raspberry Pi’s six feet apart and measuring the RSSI values.  

An AI-based contact tracing app named COVI developed in Canada leverages probabilistic risk levels to profile an individual’s infection risk level \cite{Alsdurf}. COVI uses the advantages of ML algorithms to optimize and automate the integration of pseudonymized user data in assessing the risk levels. An a priori version of an epidemiological model-based simulated dataset is used to pre-train the ML models. Upon collection of real data through an app, the simulator parameters are tuned to match with real data. The impact of ML in the COVI app is observed by using the ML predictor inside the simulator to influence the behavior of the agent in recommending the risk levels. The contact tracing application can be used to predict the lockdown area based on places visited by an infected patient. In \cite{Maghdid} the authors proposed a K-Means clustering algorithm with DASV seeding to predict the lockdown area. The proposed method has been tested in Denver, USA and successfully identified the area to be locked down as users walking in the area approach each other very frequently.  
Despite the significant advantages of using DCT systems, there are issues related to data privacy and use. However, these are out of the scope of this review paper. 

  
\begin{table}[!htbp]
\centering
\caption{ML based research studies in contact tracing.}
\label{pharma1}
\begin{tabular}{|p{0.10\linewidth}|p{0.15\linewidth}|p{0.22\linewidth}|p{0.23\linewidth}|p{0.23\linewidth}|}
\hline
\sc Study & \sc   Objective &  \sc   Methodology  &  \sc    Data & \sc  Results \\
\hline

\scriptsize He \& Printz, (2020)    &    
\scriptsize  Detect proximity of individuals. &    
\scriptsize      A 2-stage classifier that utilizes vanilla neural network to extract features from a signal emanating from different sources.
 &  \scriptsize National Institute of Standards and Technology (NIST) COVID-19 Data repository.  https://covid19-data.nist.gov/ 
 & \scriptsize No comparison, however, it shows the variation of different phone carriage states and prediction accuracy.  \\\hline

\scriptsize Shankar et al., (2020)   &  
\scriptsize Comparative study to evaluate the accuracy of Bluetooth-based distance measurement. &      
\scriptsize       Deep learning, SVM, and decision tree-based classification models are compared.
 &  \scriptsize MIT PACT Range-Angle Data set  
https://github.com/mitll/MIT- Matrix-Data and  MITRE Covid-19 synthetic data set 
 & \scriptsize Conv 1d exhibits on average 90\% better performance than other classification models.     \\\hline

 \scriptsize Song, (2020)   &  
\scriptsize Comparative study of different algorithms to determine the proximity within six feet.&      
\scriptsize       Various ML algorithms are tested in classification of Bluetooth beacon RSSI values.
 &  \scriptsize piPact, Beaver Works Summer Institute, July 2020, https://beaverworks.ll.mit.edu /CMS/bw/pipact  
 & \scriptsize Gaussian SVM shows better performance than logistic regression, k-nearest neighbors, and decision tree-based classifier with training, validation and testing accuracy of 76.12\%, 72.60\%, and 79.67\% respectively.      \\\hline

 \scriptsize Alsdurf et al., (2020)    &  
\scriptsize Capture the dependencies across the whole history of the user. &      
\scriptsize       Transformer deep learning architecture is used as the base algorithm.
 &  \scriptsize Simulated dataset based on epidemiological model.
 & \scriptsize ML-based risk prediction could reduce the reproduction number compared to standard digital contact tracing applications.  
 \\\hline

 \scriptsize Maghdid et al., (2020)   &  
\scriptsize Predict the lockdown area based on people movement. &      
\scriptsize       K-Nearest unsupervised machine learning algorithm is used for prediction.
 &  \scriptsize Android based smartphone application for user data collection.
 & \scriptsize Used a threshold of five meters the proposed protocol predicts the lockdown area.     \\\hline

\end{tabular}
\end{table}

\section{Conclusion}
Machine learning has become a potent tool in many applications. In particular, it has recently been employed in the battle against COVID-19. There exists a growing body of literature that is dedicated to the subject. The decision by the major publishers to make all COVID-19 related research publicly available has improved information flow. In this paper, we attempt to provide an overview of the rapidly increasing corpus of research in machine learning related to COVID-19. We discuss the state-of-the-art research including the material on research archives.
In particular, we covered four major areas of ML research related to COVID-19: forecasting, medical diagnostics,  drug development, and contact tracing.

Our survey revealed the following key observations.
In forecasting, recurrent neural network such as LSTMs have been used to predict the future infection and death rates. Many studies are focused on the North American region, but also other countries including Brazil and China. The best models achieve correlation of 0.999. In medical diagnostics, deep learning models that have previously shown success in other domains are being deployed to detect the presence of  the infection based on CT scans and X-rays. The best models achieve accuracy rate of 99\%. In drug discovery, a variety of algorithms are being used to develop new vaccine against the infection. However, the majority of the studies are still in the initial stage. In contact tracing, AI based applications
are utilized to identify and locate potential virus carriers though with limited success.

Despite the tremendous progress, the current machine learning approaches suffer from two major drawbacks. First, the underlying algorithms have not yet reached the level of human reasoning. The deep learning models such as CNNs, LSTMs, Transformer, and others remain imperfect and cannot consistently outperform a human expert. Second, the lack of data hinders the training and development of the models. Patient data is notoriously difficult to obtain. Since deep learning models rely on abundance of data the lack of thereof results in suboptimal generalization performance. 

Our main recommendation based on the extensive survey of current literature is the involvement of government agencies to facilitate  procurement of COVID-19 related data. Public institutions and government agencies can play a key role in obtaining and disseminating data from hospitals to researchers. Since machine learning algorithms rely heavily on large amounts of data its availability can drastically improve results.


\end{document}